\documentclass{article}

\usepackage{microtype}
\usepackage{graphicx}
\usepackage{subcaption}
\usepackage{booktabs} % for professional tables

\usepackage{hyperref}
\usepackage{placeins}

\usepackage[preprint]{icml2026}

\usepackage{amsmath}
\usepackage{amssymb}
\usepackage{mathtools}
\usepackage{amsthm}
\usepackage{float}
\usepackage{algorithm}
\usepackage{algorithmic}
\usepackage{multirow}
\usepackage[capitalize,noabbrev]{cleveref}

\theoremstyle{plain}

\theoremstyle{definition}

\theoremstyle{remark}

\usepackage[textsize=tiny]{todonotes}

\icmltitlerunning{EDIS: Diagnosing LLM Reasoning via Entropy Dynamics}
\PassOptionsToPackage{hyphens}{url}

\usepackage{xurl}      % 比 \usepackage{url} 更能断行
\usepackage{hyperref}  % 尽量放最后
\hypersetup{breaklinks=true}

\begin{document}

% EDIS: A Trajectory-Level Metric of Instability in Entropy Evolution

\twocolumn[
  \icmltitle{EDIS: Diagnosing LLM Reasoning via Entropy Dynamics}
  \icmlsetsymbol{equal}{*}

  \begin{icmlauthorlist}
    \icmlauthor{Chenghua Zhu}{equal,scnu}
    \icmlauthor{Siyan Wu}{equal,scnu}
    \icmlauthor{Xiangkang Zeng}{sysu}
    \icmlauthor{Zishan Xu}{sjtu}
    \icmlauthor{Zhaolu Kang}{pku}
    \icmlauthor{Yifu Guo}{sysu}
    \icmlauthor{Yuquan Lu}{scnu}
    \icmlauthor{Junduan Huang}{scnu}
    \icmlauthor{Guojing Zhou}{scnu}
  \end{icmlauthorlist}

  \icmlaffiliation{scnu}{South China Normal University, Guangzhou, China}
  \icmlaffiliation{sysu}{Sun Yat-sen University, Guangzhou, China}
  \icmlaffiliation{sjtu}{Shanghai Jiao Tong University, Shanghai, China}
  \icmlaffiliation{pku}{Peking University, Beijing, China}

  \icmlcorrespondingauthor{Guojing Zhou}{gjzhou86@m.scnu.edu.cn}

  \icmlkeywords{Large Language Models, Reasoning, Entropy Dynamics, Uncertainty, Token-level Analysis, Best-of-N Selection, Reinforcement Learning}

  \vskip 0.135in
]



\printAffiliationsAndNotice{\icmlEqualContribution}

\begin{abstract}
Entropy-based confidence signals are increasingly leveraged to improve reasoning in large language models (LLMs), yet existing approaches treat confidence as a static quantity---typically aggregated over tokens. We show that the \emph{temporal evolution} of confidence during generation carries richer information than aggregate statistics alone. Analyzing token-level entropy trajectories, we identify characteristic patterns distinguishing correct from incorrect reasoning: erroneous solutions exhibit unstable dynamics, including burst spikes (sustained uncertainty growth) and peak-valley spikes (sharp rebounds following transient confidence). These patterns persist across models and training stages, suggesting they reflect intrinsic properties of reasoning failure rather than superficial noise. To formalize this observation, we introduce the Entropy Dynamics Instability Score (\textbf{EDIS}), a trajectory-level metric quantifying instability in entropy evolution. EDIS serves as an effective diagnostic signal for inference-time selection, substantially improving reasoning accuracy, and offers a promising direction for training-time sample curation. Our findings establish entropy dynamics as an underexplored yet informative lens for understanding and improving LLM reasoning.
\end{abstract}

% TL; DR: We propose EDIS (Entropy Dynamics Instability Score), which captures how entropy evolves during LLM generation to distinguish correct from incorrect reasoning, achieving 82% accuracy improvement for selection and +7.7 pp gains for RL training.

\section{Introduction}

Large language models (LLMs) have achieved remarkable progress on complex reasoning tasks~\cite{wei2022cot,wang2023selfconsistency}, yet a fundamental challenge persists: distinguishing correct reasoning from plausible-sounding errors remains difficult without external verification~\cite{farquhar2024semanticentropy,kapoor2024taughtknow}. A natural approach is to leverage the model's own confidence signals---typically entropy or token-level probabilities---to identify unreliable outputs~\cite{guo2017calibration,chen2024quantifying}. However, existing methods treat confidence as a \emph{static} quantity, aggregating token-level uncertainty into summary statistics or examining only the final output. Recent evidence suggests that entropy calibration degrades during autoregressive generation~\cite{cao2025entropycalibration}, indicating that this static view may miss important structure. More fundamentally, it overlooks a key aspect of autoregressive generation: reasoning unfolds sequentially, and confidence evolves throughout the process.

\begin{figure*}[thb]
\centering
\includegraphics[width=0.75\textwidth]{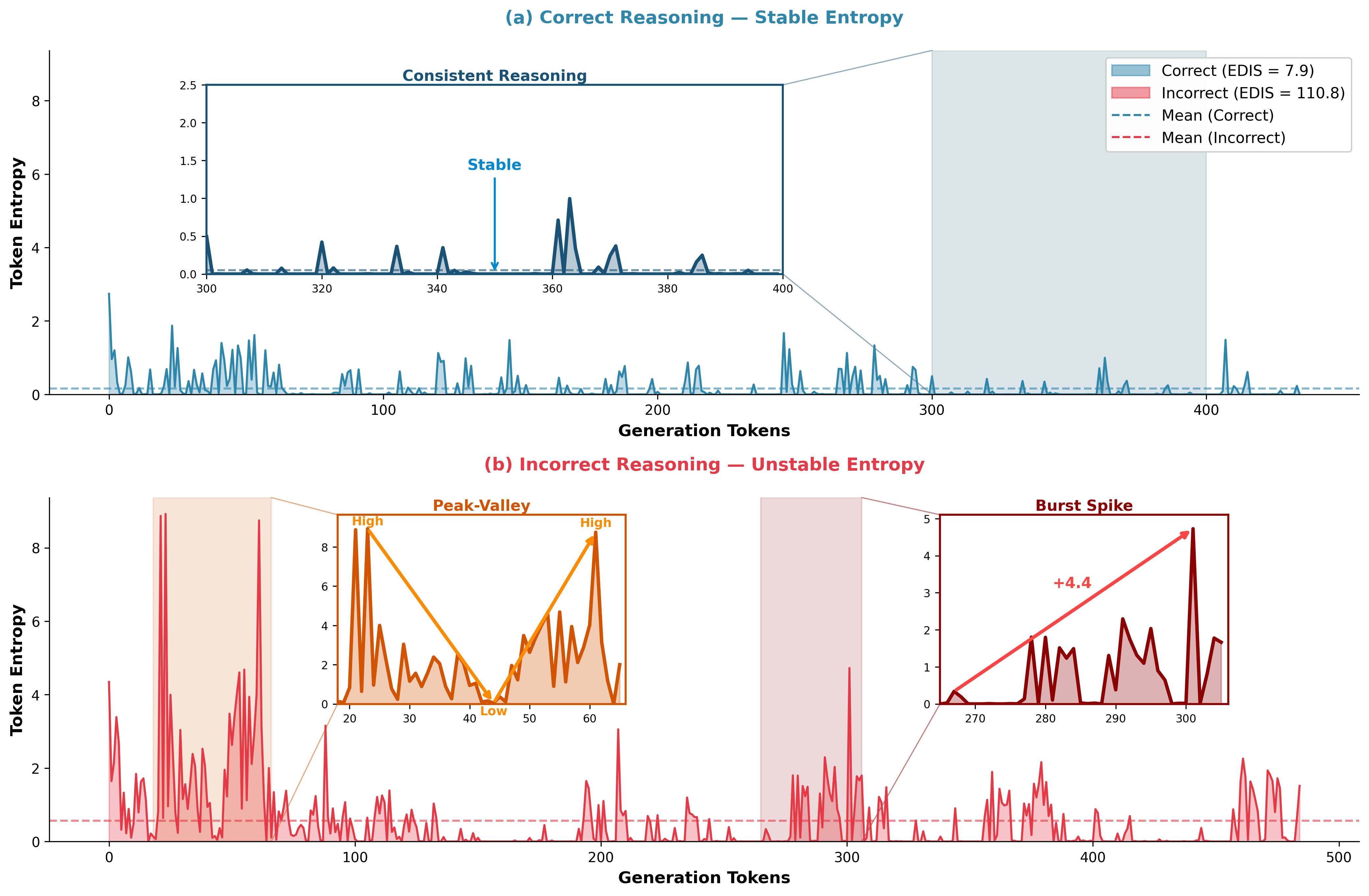}
\caption{Token entropy trajectories for correct (top) and incorrect (bottom) reasoning. Correct responses maintain stable, low entropy, while incorrect responses exhibit distinctive instability patterns: peak-valley spikes (entropy drops then rebounds) and burst spikes (progressive entropy rise).}
\label{fig:entropy_patterns}
\end{figure*}

In this work, we demonstrate that \textbf{how confidence evolves during generation is more informative than its aggregate value}. Through systematic analysis of token-level entropy trajectories, we uncover a striking pattern: incorrect reasoning is not merely associated with higher uncertainty, but with \emph{instability} in how uncertainty evolves. As illustrated in Figure~\ref{fig:entropy_patterns}, correct reasoning produces relatively smooth entropy curves where most tokens exhibit low entropy with few spikes or oscillations. In contrast, incorrect reasoning shows frequent high-entropy tokens and characteristic instability patterns. We identify two typical failure modes: \emph{burst spikes}, where entropy rises steadily over consecutive tokens as the model becomes progressively confused, and \emph{peak-valley (rebound) spikes}, where entropy drops to a local minimum before sharply rebounding---indicating false confidence followed by renewed uncertainty. These instability patterns are remarkably consistent: across models, temperatures, and training stages, incorrect responses exhibit $1.7$--$3.6\times$ more entropy fluctuations than correct ones (Cohen's $d \approx 1.0$), suggesting they reflect fundamental properties of reasoning failure rather than incidental noise.

\begin{figure}[t]
\centering
\includegraphics[width=\columnwidth]{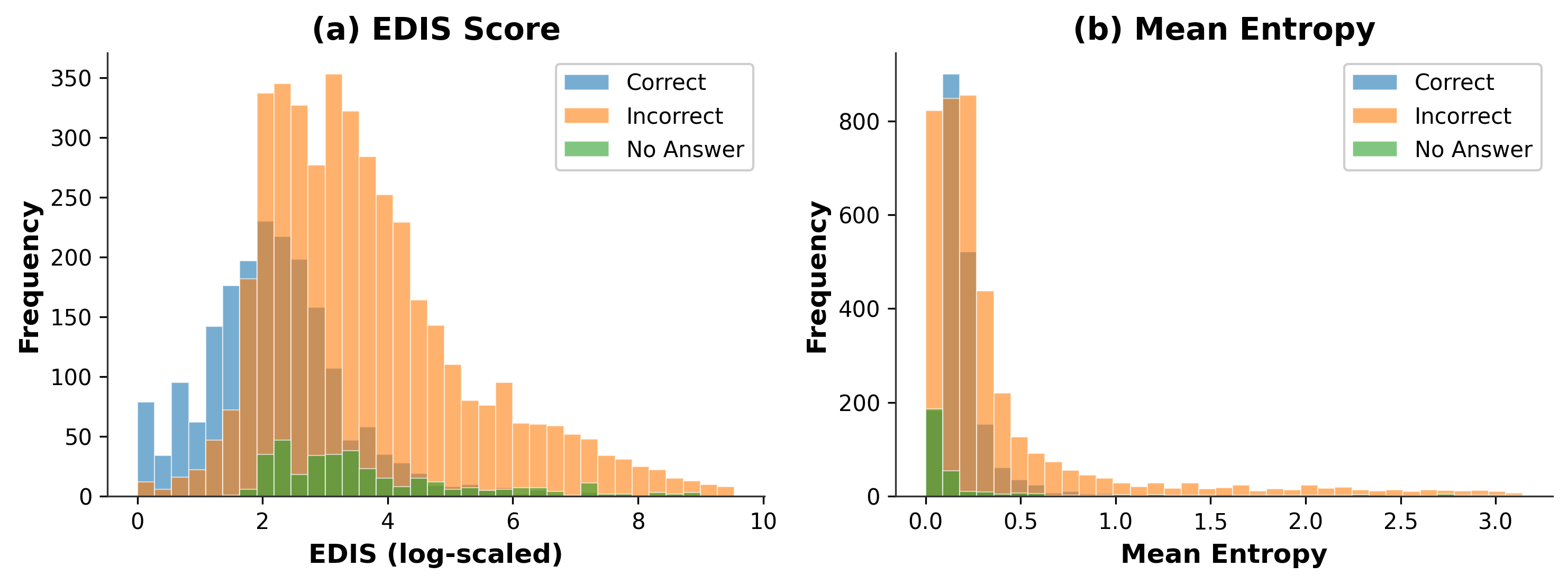}
\caption{EDIS (left) vs.\ mean entropy (right) distributions. EDIS clearly separates correct from incorrect responses, while mean entropy distributions largely overlap.}
\label{fig:edis_distribution}
\vspace{-0.3cm}
\end{figure}

To operationalize this observation, we introduce the \emph{Entropy Dynamics Instability Score (\textbf{EDIS})}, a simple trajectory-level metric that captures two complementary forms of instability: \emph{burst spikes} (cumulative entropy growth within a sliding window) and \emph{peak-valley spikes} (sharp increases from historical minima). As shown in Figure~\ref{fig:edis_distribution}, EDIS distributions for correct and incorrect responses concentrate around distinct central values, enabling clear separation. In contrast, mean entropy---a common baseline---fails to distinguish response quality, highlighting the value of trajectory-level analysis.

We validate EDIS through extensive experiments on mathematical reasoning. For inference-time selection, EDIS-based filtering substantially improves answer quality: across four benchmarks and three models, average accuracy improves from $29.9\%$ to $54.5\%$---an $82\%$ relative gain---without requiring verifiers or additional annotations. Compared with alternative confidence measures, EDIS achieves $60.6\%$ overall accuracy versus $51.7\%$ for self-certainty and $50.9\%$ for sequence entropy. We also present preliminary evidence that EDIS can inform training-time sample curation in reinforcement learning. These results establish entropy dynamics as an informative signal for assessing reasoning quality that complements static confidence measures.

Our contributions are as follows:
\vspace{-0.3cm}
\begin{itemize}
    \item We conduct a systematic empirical analysis of entropy dynamics in LLM reasoning, revealing that incorrect solutions exhibit characteristic instability patterns---burst spikes and peak-valley spikes---that persist across models and training stages.
    \item We introduce EDIS, a simple and interpretable trajectory-level metric that quantifies entropy instability by combining burst spike detection (cumulative growth) and peak-valley spike detection (deviation from historical minima).
    \item We validate EDIS effectiveness through extensive experiments: EDIS-based selection achieves $82\%$ relative improvement in accuracy and consistently outperforms alternative confidence measures. We also present preliminary evidence that EDIS can inform training-time sample curation.
\end{itemize}

\section{Related Work}
\label{sec:related}

Entropy and confidence signals have been extensively studied in language models, yet existing methods share a common limitation: they treat confidence as a \emph{static} quantity, collapsing the generation process into summary statistics. We review prior work across three areas and identify the gap that motivates our trajectory-level approach.

\paragraph{Uncertainty and Confidence Estimation.}
Quantifying model confidence is a long-standing challenge in machine learning~\cite{guo2017calibration,desai2020calibration}. In language models, common approaches aggregate token-level probabilities or entropy into sequence-level scores for calibration and selective prediction~\cite{kadavath2022languagemodels}. The simplest approach computes mean entropy $\bar{H} = \frac{1}{T}\sum_{t=1}^{T} H_t$ across tokens, capturing average uncertainty but discarding temporal information. Semantic entropy~\cite{farquhar2024semanticentropy} groups semantically equivalent outputs to reduce spurious variability, enabling meaning-aware hallucination detection, but still produces a single scalar per generation. Self-certainty~\cite{kang2025selfcertainty} measures probability mass concentration in the distribution tail, providing a token-level confidence signal. Recent work on entropy minimization~\cite{agarwal2025rent,zhao2025learningreason} demonstrates that lower entropy correlates with reasoning accuracy, with tokens near final answers exhibiting the strongest signal. However, all these methods collapse the generation process into summary statistics---averaging entropy or examining only specific tokens---discarding the temporal structure of how confidence evolves throughout reasoning.

\paragraph{Inference-Time Scaling.}
Scaling compute at inference time has emerged as an effective strategy for improving reasoning~\cite{snell2024scaling,brown2024large}. Common approaches generate multiple candidates and select among them via majority voting~\cite{wang2023selfconsistency}, verifier-based reranking~\cite{cobbe2021trainingverifiers,lightman2024verify}, or confidence-based filtering. Entropy signals have been applied to early stopping in chain-of-thought reasoning~\cite{thinkjustenough2025} and selective abstention~\cite{ren2023selfeval}. Yet these methods assess reliability from aggregate confidence scores, implicitly assuming that a single scalar suffices to characterize reasoning quality. Process-aware verification validates intermediate steps~\cite{lightman2024verify} but requires external verifiers or human annotations, limiting scalability.

\paragraph{Entropy Signals in Training.}
Entropy also plays a critical role in reinforcement learning for LLM reasoning. Policy entropy collapse---where models become excessively deterministic during training---limits performance scaling~\cite{cui2025entropymechanism}, motivating work on entropy-aware objectives such as maximum entropy RL~\cite{haarnoja2018soft}. Complementary work defines intrinsic rewards based on entropy or confidence, enabling unsupervised reasoning without external supervision: RENT~\cite{agarwal2025rent} trains models to minimize token-level entropy, while \citet{zhao2025learningreason} rewards self-certainty maximization. These approaches directly modify the training objective by using entropy or confidence as the reward. In contrast, EDIS preserves the original reward signal and uses entropy dynamics to \emph{curate} training data---strengthening more informative sequences and down-weighting less informative ones.

\section{Preliminaries}

\begin{figure*}[h]
\centering
\includegraphics[width=\textwidth]{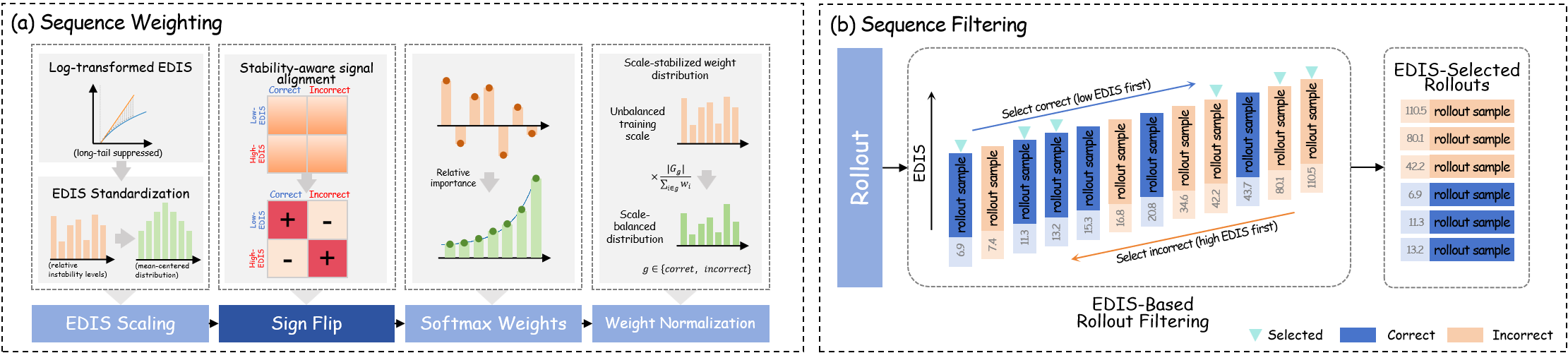}
\caption{EDIS-based sample curation for RL training.}
\label{fig:rl_pipeline}
\end{figure*}

We begin by establishing the formal framework for analyzing entropy dynamics in autoregressive language models and introduce the reinforcement learning algorithm used in our experiments.

\paragraph{Token-Level Entropy.}
Consider an autoregressive language model parameterized by $\theta$ that generates a response $\mathbf{y} = (y_1, y_2, \ldots, y_T)$ conditioned on a prompt $\mathbf{x}$. At each generation step $t$, the model produces a distribution $\pi_\theta(y_t \mid \mathbf{x}, y_{<t})$ over the vocabulary $\mathcal{V}$. We quantify the model's uncertainty at position $t$ via the \emph{token entropy}:
\begin{equation}
H_t = -\sum_{v \in \mathcal{V}} \pi_\theta(v \mid \mathbf{x}, y_{<t}) \log \pi_\theta(v \mid \mathbf{x}, y_{<t}).
\label{eq:token_entropy}
\end{equation}
Low entropy indicates high confidence, with probability mass concentrated on a small set of tokens; high entropy reflects uncertainty, with probability spread more uniformly across the vocabulary.

\paragraph{Entropy Trajectory.}
While prior work typically aggregates token-level entropy into scalar summaries (see Section~\ref{sec:related}), we argue that the \emph{temporal structure} of entropy evolution carries important information. We define the \emph{entropy trajectory} as the ordered sequence:
\begin{equation}
\mathbf{H} = (H_1, H_2, \ldots, H_T),
\label{eq:entropy_trajectory}
\end{equation}
where each $H_t$ is computed according to Eq.~\eqref{eq:token_entropy}. This representation preserves the dynamics of uncertainty throughout generation, enabling analysis of \emph{how} confidence evolves rather than merely \emph{how confident} the model is on average. As we demonstrate empirically, this trajectory-level perspective reveals diagnostic patterns that distinguish correct from incorrect reasoning.

\paragraph{Group Relative Policy Optimization.}
We adopt Group Relative Policy Optimization (GRPO)~\cite{shao2024grpo} as our RL training framework. For each prompt $\mathbf{x}$, GRPO samples a group of $G$ responses $\{\mathbf{y}^{(1)}, \ldots, \mathbf{y}^{(G)}\}$ from the old policy $\pi_{\theta_{\text{old}}}$ and computes a reward $r^{(i)}$ for each. The objective maximizes a clipped surrogate with group-relative advantages:
\begin{equation}
\mathcal{J}_{\text{GRPO}}(\theta) = \mathbb{E} \left[ \frac{1}{G} \sum_{i=1}^{G} \min\bigl( \rho^{(i)} \hat{A}^{(i)}, \, \bar{\rho}^{(i)} \hat{A}^{(i)} \bigr) \right],
\label{eq:grpo_objective}
\end{equation}
where $\rho^{(i)} = \pi_\theta(\mathbf{y}^{(i)}|\mathbf{x}) / \pi_{\theta_{\text{old}}}(\mathbf{y}^{(i)}|\mathbf{x})$ is the importance ratio, $\bar{\rho}^{(i)} = \text{clip}(\rho^{(i)}, 1-\epsilon, 1+\epsilon)$ is the clipped ratio, and the advantage is estimated relative to the group:
\begin{equation}
\hat{A}^{(i)} = \frac{r^{(i)} - \mu_G}{\sigma_G},
\label{eq:grpo_advantage}
\end{equation}
with $\mu_G$ and $\sigma_G$ being the mean and standard deviation of rewards within the group. This eliminates the need for a separate critic network while providing stable advantage estimates through within-group normalization.

\section{Methods - Entropy Dynamics Instability Score (EDIS)}
\label{sec:edis}

We first describe the empirical phenomena that motivate our approach, then present the formal definition of the Entropy Dynamics Instability Score (\textbf{EDIS}).

\subsection{Characteristic Instability Patterns}
\label{sec:instability_patterns}

Analyzing rollouts from Qwen2.5-Math-1.5B across training checkpoints and temperatures reveals that incorrect solutions exhibit significantly more entropy spikes than correct ones ($1.7$--$3.6\times$ more; Appendix~\ref{app:spike_analysis}). Beyond aggregate counts, we identify two characteristic temporal patterns that reliably distinguish incorrect reasoning:

\paragraph{Burst Spikes.}
A recurring signature of incorrect reasoning is \emph{sustained entropy increase} over consecutive tokens (Figure~\ref{fig:entropy_patterns}, bottom-right). Rather than a single abrupt jump, entropy rises steadily across a window of $w$ tokens, indicating progressive deterioration of model confidence. We formalize this pattern by counting positions where cumulative entropy growth exceeds a threshold:
\begin{equation}
S_{\text{burst}} = \sum_{t=1}^{T-w} \mathbb{I}\bigl(H_{t+w} - H_t > \tau_b\bigr),
\label{eq:burst_spike}
\end{equation}
where $\mathbb{I}(\cdot)$ is the indicator function, $w$ is the window size, and $\tau_b$ is the threshold for detecting significant entropy growth. This captures the intuition that ``the more the model generates, the more confused it becomes''---a hallmark of reasoning gone astray.

\paragraph{Peak-Valley Spikes.}
A second pattern involves \emph{false confidence followed by renewed uncertainty}---a characteristic V-shaped trajectory (Figure~\ref{fig:entropy_patterns}, bottom-left). The running minimum $\min_{s < t} H_s$ tracks the most confident state the model has achieved so far; when current entropy $H_t$ rises significantly above this baseline, it indicates that previously attained confidence has eroded. We count such events as:
\begin{equation}
S_{\text{rebound}} = \sum_{t=2}^{T} \mathbb{I}\bigl(H_t - \min_{s < t} H_s > \tau_r\bigr),
\label{eq:rebound_spike}
\end{equation}
where $\tau_r$ is the threshold for detecting significant rebound from historical minima. This formulation inherently captures V-shaped dynamics: the running minimum only becomes low if entropy has previously \emph{decreased} (the descent into the valley), and the threshold is only exceeded when entropy subsequently \emph{rises} (the ascent out of it). The pattern signals that the model reached a confident state but then encountered renewed difficulty.

% \vspace{0.3em}
\noindent Both patterns are qualitatively distinct from correct reasoning, which exhibits smoother entropy evolution with fewer abrupt transitions. Appendix~\ref{appendix:token_heatmap} provides token-level visualizations illustrating these dynamics.

\subsection{Metric Definition}
\label{sec:edis_definition}

The instability patterns described above suggest that reasoning quality can be diagnosed from entropy trajectory structure. We operationalize this insight with the \emph{Entropy Dynamics Instability Score (\textbf{EDIS})}, a trajectory-level metric that combines spike frequency with overall variance:
\begin{equation}
\mathrm{EDIS}(\mathbf{H}) = S(\mathbf{H}) \cdot \bigl(1 + \mathrm{Var}(\mathbf{H})\bigr),
\label{eq:edis}
\end{equation}
where $S(\mathbf{H}) = \tfrac{1}{2}(S_{\text{burst}} + S_{\text{rebound}})$ denotes the combined spike score and
\begin{equation}
\mathrm{Var}(\mathbf{H}) = \frac{1}{T}\sum_{t=1}^{T}(H_t - \bar{H})^2, \quad \bar{H} = \frac{1}{T}\sum_{t=1}^{T} H_t
\label{eq:entropy_var}
\end{equation}
is the entropy variance. The multiplicative formulation captures the intuition that reasoning is most unstable when spike events co-occur with high overall variance. Lower EDIS indicates more stable reasoning; $\mathrm{EDIS} \approx 0$ corresponds to smooth, confident generation.

\subsection{EDIS for Reinforcement Learning}
\label{sec:edis_rl}

While the primary application of EDIS is inference-time selection, the same principle---that entropy stability indicates reasoning quality---may also inform training. We present a preliminary exploration of using EDIS to curate training samples in RL, as a proof-of-concept rather than a fully optimized approach.

The intuition is straightforward: not all training samples are equally informative. Confident correct responses (low EDIS) represent reliable reasoning worth reinforcing, while struggling incorrect ones (high EDIS) reveal genuine difficulties worth learning from. In contrast, lucky correct guesses (high EDIS) and systematic failures (low EDIS) provide weaker learning signals. We explore two mechanisms that leverage this structure (Figure~\ref{fig:rl_pipeline}): \emph{sequence filtering} retains only high-signal trajectories, and \emph{sequence weighting} assigns differential importance to all samples.

\paragraph{Sequence Filtering.}
The simplest approach retains only extreme trajectories: the most stable correct responses (lowest EDIS) and the most unstable incorrect ones (highest EDIS). In practice, we oversample $m \cdot n$ candidates per prompt and alternately select from these extremes until $n$ samples remain, discarding ambiguous cases entirely.

\paragraph{Sequence Weighting.}

Rather than discarding samples, we can assign differential weights based on EDIS. Since raw EDIS scores often exhibit skewed distributions with long tails, we apply a log transformation to compress the range and then standardize to z-scores: $z_i = (\log(\mathrm{EDIS}_i + 1) - \mu) / \sigma$, where $\mu$ and $\sigma$ are batch statistics. This normalization ensures consistent weighting across batches with different EDIS distributions. After standardization, $z_i > 0$ indicates above-average instability and $z_i < 0$ indicates below-average instability. A correctness-dependent transformation ensures informative samples receive higher weights:
\begin{equation}
s_i = \begin{cases}
-z_i & \text{if correct} \\
\phantom{-}z_i & \text{if incorrect}
\end{cases}
\label{eq:signed_edis}
\end{equation}
For correct trajectories, we negate to favor stability (low EDIS $\to$ high weight); for incorrect trajectories, we preserve the sign to favor instability (high EDIS $\to$ high weight). Weights are computed via $w_i = \mathrm{softmax}(s_i / \alpha) \cdot n$, where $\alpha$ controls concentration. To preserve gradient balance, we normalize separately within the correct and incorrect groups:
\begin{equation}
w_i^{\text{norm}} = \frac{w_i}{\sum_{j \in \mathcal{G}_i} w_j} \cdot |\mathcal{G}_i|,
\label{eq:group_norm}
\end{equation}
where $\mathcal{G}_i$ denotes the correctness group (correct or incorrect) containing sample $i$. The weighted advantage becomes $\tilde{A}_i = A_i \cdot w_i^{\text{norm}}$. We apply weighting only to prompts with mixed outcomes.

\section{Experiments}
\label{sec:experiments}

This section evaluates EDIS in two settings: inference-time selection (Sections~\ref{sec:edis_scaling}--\ref{sec:edis_entropy}) and reinforcement learning (Section~\ref{sec:edis_rl_exp}).

\subsection{Best-of-$N$ Selection}
\label{sec:edis_scaling}

The first question is whether EDIS captures meaningful signal for inference-time selection. Unlike verifier-based methods that require additional training or supervision, EDIS leverages only the model's internal uncertainty dynamics. If EDIS reliably distinguishes response quality, filtering from larger candidate pools should yield systematic accuracy gains.

\begin{figure*}[tb]
  \centering
  \includegraphics[width=\textwidth]{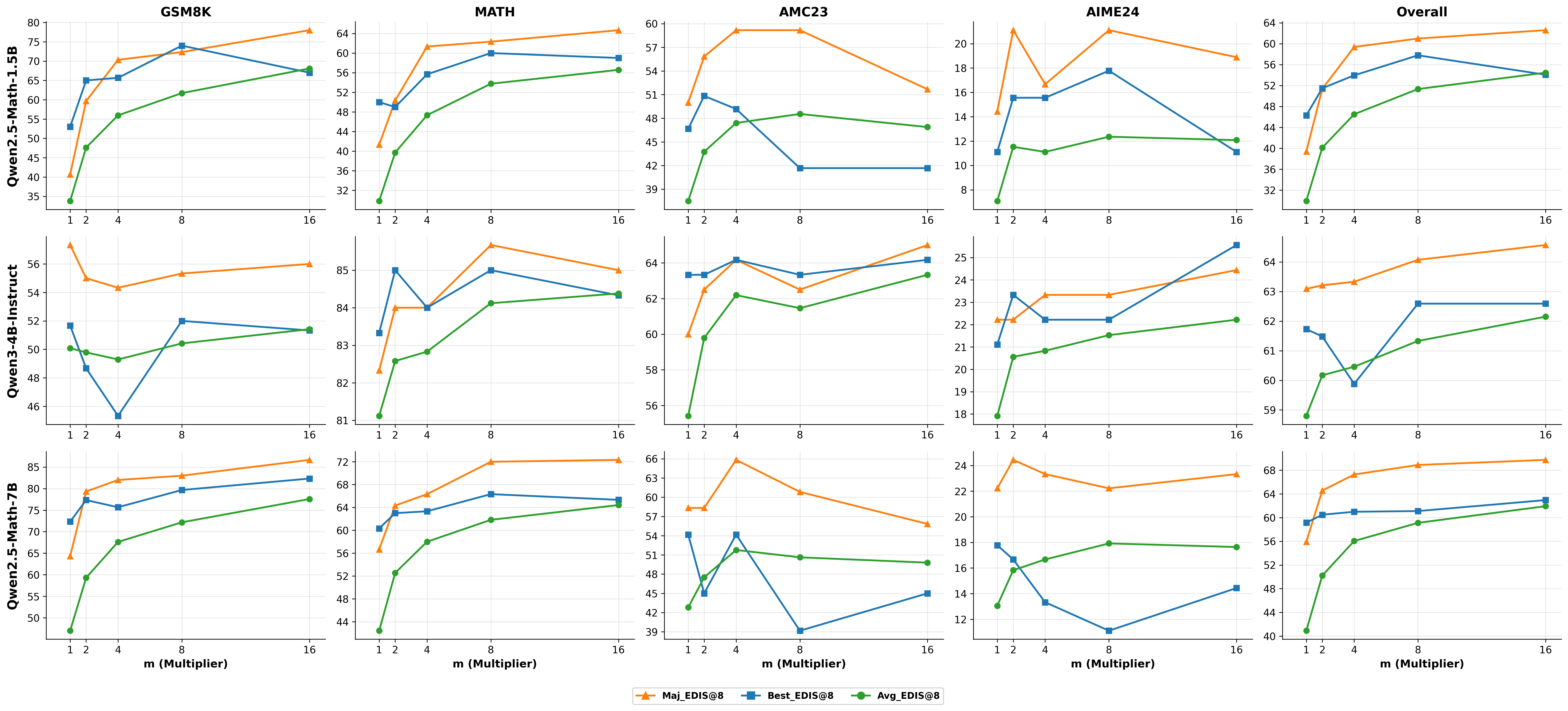}
  \caption{EDIS-based best-$k$-of-$N$ selection across models and benchmarks ($k = 8$). Accuracy improves consistently as the oversampling multiplier $m$ increases. Results averaged across three temperatures ($0.2$, $0.6$, $1.0$).}
  \label{fig:best_of_n}
\end{figure*}

\paragraph{Setting.} Experiments span four mathematical reasoning benchmarks---GSM8K~\cite{cobbe2021trainingverifiers}, MATH~\cite{hendrycks2021math}, AMC23~\cite{knoveleng_amc23}, and AIME24~\cite{huggingfaceh4_aime2024}---using three models: Qwen2.5-Math-1.5B~\cite{qwen25math15b_model,qwen25math_report}, Qwen3-4B-Instruct~\cite{qwen3_4b_instruct_model,qwen3_report}, and Qwen2.5-Math-7B~\cite{qwen25math7b_model,qwen25math_report}. For GSM8K and MATH, we randomly sample 100 problems; for AMC23 and AIME24, we use the full test sets. All experiments use three sampling temperatures ($0.2$, $0.6$, $1.0$), with results averaged across temperatures. For each problem, we generate $N = m \cdot k$ candidates ($k = 8$, $m \in \{1, 2, 4, 8, 16\}$), rank by EDIS, and retain the $k$ most stable responses (lowest EDIS).

\paragraph{Results.} Figure~\ref{fig:best_of_n} shows remarkably consistent improvements: across all three models, four benchmarks, and three metrics (average accuracy, EDIS-best, and majority voting), accuracy increases monotonically with the oversampling multiplier. The gains are substantial, particularly for models with lower baseline performance. Aggregating across benchmarks, Qwen2.5-Math-1.5B improves from $29.9\%$ to $54.5\%$ in average accuracy as $m$ increases from $1$ to $16$---a gain of $24.6$ percentage points that nearly doubles the baseline. Similarly, Qwen2.5-Math-7B improves from $40.9\%$ to $61.9\%$ (+$21.0$ pp). Even the stronger Qwen3-4B-Instruct, starting at $58.8\%$, achieves consistent gains to $62.2\%$ (+$3.4$ pp). The pattern holds for majority voting, where Qwen2.5-Math-1.5B improves from $39.4\%$ to $62.6\%$ (+$23.2$ pp) and Qwen2.5-Math-7B from $55.9\%$ to $69.8\%$ (+$13.9$ pp). Notably, even EDIS-best accuracy improves (e.g., $46.3\%$ to $54.1\%$ for Qwen2.5-Math-1.5B), indicating that EDIS filtering enriches the candidate pool with higher-quality responses rather than merely improving answer aggregation. These results confirm that EDIS reliably indicates reasoning quality, enabling substantial improvements without external supervision. Detailed per-model breakdowns across all temperatures are provided in Appendix~\ref{app:best_of_n_detail}.

\subsection{Comparison with Other Selection Methods}
\label{sec:edis_comparison}

To evaluate the effectiveness of EDIS relative to other selection methods, we compare against several baselines: \textbf{Mean} (unweighted average), \textbf{Majority Voting} (most frequent answer), \textbf{Sequence Entropy} (mean token-level entropy $\bar{H} = \frac{1}{T}\sum_{t=1}^{T} H_t$, lower indicates higher confidence), and \textbf{Self-Certainty (SC)}~\cite{kang2025selfcertainty}, which measures confidence via KL divergence between the predicted distribution and a uniform distribution over the vocabulary, aggregated across tokens; higher values indicate the model concentrates probability mass on fewer tokens, suggesting greater certainty.

\paragraph{Setting.} For confidence-based methods (Entropy, SC, EDIS), we use Score-Weighted Borda aggregation: each response casts a weighted vote for its predicted answer, with the highest-weighted answer selected. For metrics where lower values indicate higher confidence (Entropy, EDIS), we use inverse weighting $w_i = (s_i + \epsilon)^{-1}$ with $\epsilon = 0.1$; for SC, we use $w_i = s_i$ directly. Experiments use Qwen2.5-Math-1.5B across four benchmarks, generating $m \in \{4, 8, 16\}$ candidates per problem, with results averaged across three temperatures.

\begin{table}[!h]
\centering
\caption{Comparison of selection methods.}
\label{tab:scaling_comparison}
\resizebox{\columnwidth}{!}{%
\begin{tabular}{cl|ccccc}
\toprule
$m$ & Dataset & Mean & Maj Vote & Entropy & SC & EDIS \\
\midrule
\multirow{5}{*}{4}
 & GSM8K & 36.0 & 47.3 & 53.3 & 56.3 & \textbf{67.3} \\
 & MATH & 30.0 & 46.3 & 50.7 & 53.0 & \textbf{58.0} \\
 & AMC23 & 38.9 & 56.7 & \textbf{58.3} & 56.7 & 55.8 \\
 & AIME24 & 7.2 & 16.7 & 18.9 & 16.7 & \textbf{21.1} \\
 & \textit{Overall} & 31.0 & 44.9 & 49.3 & 50.7 & \textbf{57.0} \\
\midrule
\multirow{5}{*}{8}
 & GSM8K & 36.3 & 49.7 & 56.7 & 58.7 & \textbf{72.3} \\
 & MATH & 29.5 & 46.0 & 50.3 & 52.3 & \textbf{60.0} \\
 & AMC23 & 38.2 & 59.2 & \textbf{60.0} & 59.2 & 57.5 \\
 & AIME24 & 7.5 & 16.7 & 16.7 & \textbf{17.8} & \textbf{17.8} \\
 & \textit{Overall} & 30.8 & 46.1 & 50.4 & 51.9 & \textbf{59.5} \\
\midrule
\multirow{5}{*}{16}
 & GSM8K & 35.6 & 49.0 & 55.0 & 56.0 & \textbf{72.3} \\
 & MATH & 29.5 & 47.3 & 52.0 & 54.0 & \textbf{62.3} \\
 & AMC23 & 37.6 & 57.5 & \textbf{60.8} & 60.0 & 55.8 \\
 & AIME24 & 6.9 & 17.8 & 20.0 & 18.9 & \textbf{22.2} \\
 & \textit{Overall} & 30.4 & 46.2 & 50.9 & 51.7 & \textbf{60.6} \\
\bottomrule
\end{tabular}%
}
\end{table}

\paragraph{Results.} Table~\ref{tab:scaling_comparison} shows that EDIS consistently achieves the highest overall accuracy across all candidate pool sizes: $57.0\%$ at $m=4$, $59.5\%$ at $m=8$, and $60.6\%$ at $m=16$---outperforming the next-best method (SC) by $6.3$ to $8.9$ percentage points. The gains are particularly pronounced on GSM8K ($72.3\%$ vs.\ $56.0\%$ for SC at $m=16$) and MATH ($62.3\%$ vs.\ $54.0\%$).

Notably, EDIS's advantage over Sequence Entropy ($60.6\%$ vs.\ $50.9\%$) shows that \emph{how} entropy evolves matters more than its average value. EDIS achieves best or tied-best performance on $9$ of $12$ dataset-$m$ combinations; Sequence Entropy occasionally wins on AMC23, where the limited test set size may reduce statistical power. These results demonstrate that EDIS provides a reliable confidence signal that scales effectively with increased inference compute. Full results broken down by temperature are reported in Appendix~\ref{app:full_scaling}.

\subsection{EDIS vs.\ Mean Entropy}
\label{sec:edis_entropy}

To further study the predictive power of EDIS, this section compares it against mean entropy at the individual sequence level. The analysis covers 26,356 valid responses (excluding no-answer responses) from Qwen2.5-Math-1.5B across four benchmarks at three temperatures.

\begin{figure}[tb]
  \centering
  \includegraphics[width=\columnwidth]{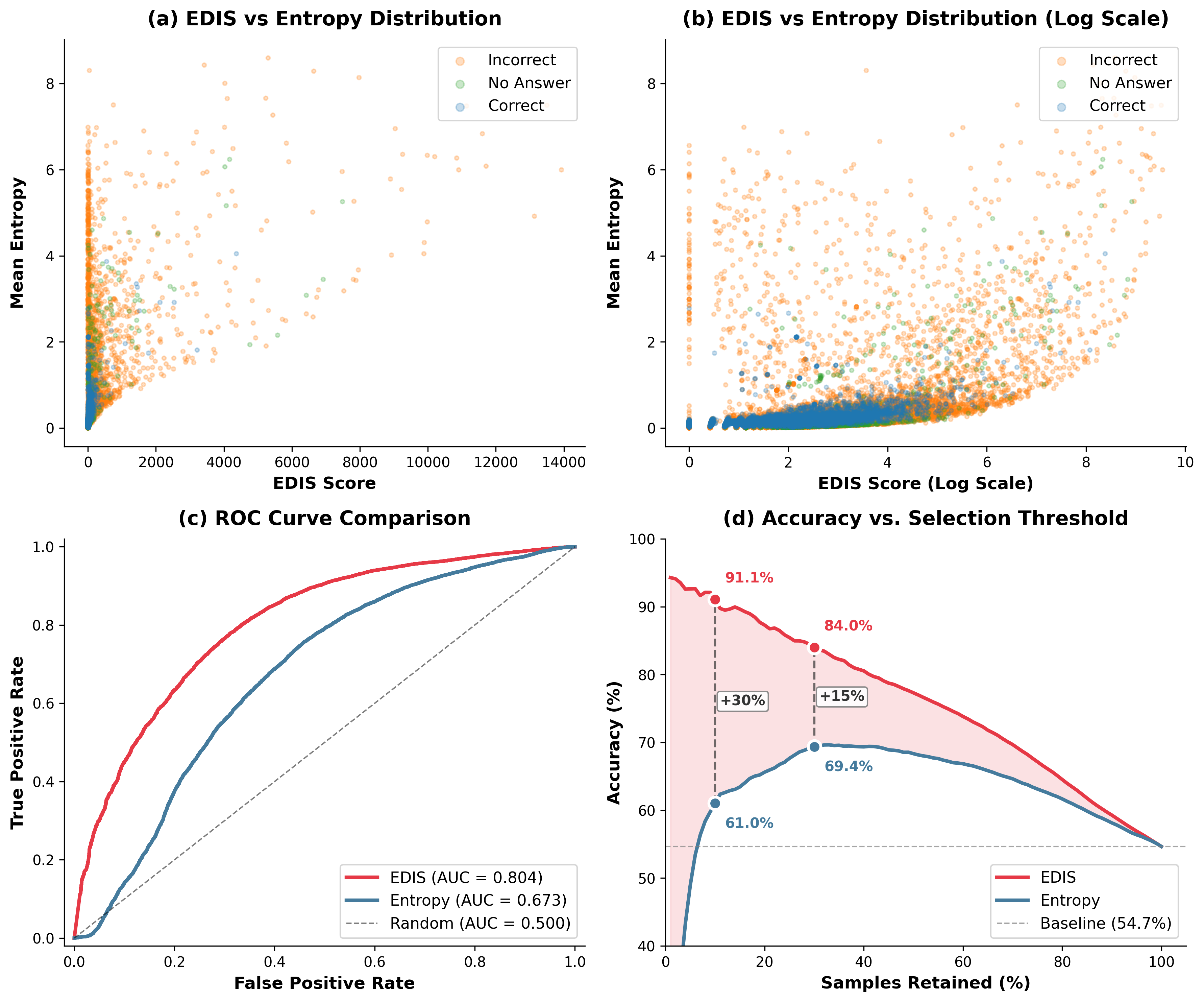}
  \caption{EDIS vs.\ Mean Entropy as correctness predictors. (a-b) EDIS provides clearer separation. (c) EDIS achieves higher AUC (0.804 vs.\ 0.673). (d) EDIS maintains advantage across retention rates.}
  \label{fig:edis_analysis}
\end{figure}

\paragraph{Correlation Analysis.}
EDIS and mean entropy are moderately correlated (Pearson $r = 0.58$, Spearman $\rho = 0.66$), indicating they capture related but distinct information. Critically, their correlations with correctness differ: mean entropy shows higher linear correlation (Pearson $r = -0.19$ vs.\ $-0.10$), but EDIS achieves substantially stronger rank correlation (Spearman $\rho = -0.52$ vs.\ $-0.30$). This reveals that EDIS captures non-linear ranking relationships crucial for selection tasks.

\paragraph{ROC-AUC Performance.}
Figure~\ref{fig:edis_analysis}(c) shows ROC curves. EDIS achieves an AUC of $0.804$---correctly ranking a random correct-incorrect pair 80.4\% of the time---compared to $0.673$ for mean entropy, a gap of $13.1$ points. The EDIS curve rises steeply at low false positive rates, indicating effective separation at aggressive thresholds.

\paragraph{Selection Accuracy.}
When selecting responses by lowest score (highest confidence), EDIS maintains consistent advantages (Figure~\ref{fig:edis_analysis}(d)). At top 10\% retention, EDIS achieves 91.1\% accuracy versus 61.0\% for entropy---a 30-point gap. The advantage persists at 20\% (+21.7 pp), 30\% (+14.6 pp), and 50\% (+9.0 pp) retention, demonstrating that trajectory dynamics capture quality signals that aggregate statistics miss.

\subsection{EDIS for Reinforcement Learning}
\label{sec:edis_rl_exp}

\begin{figure*}[!t]
\centering
\includegraphics[width=\textwidth]{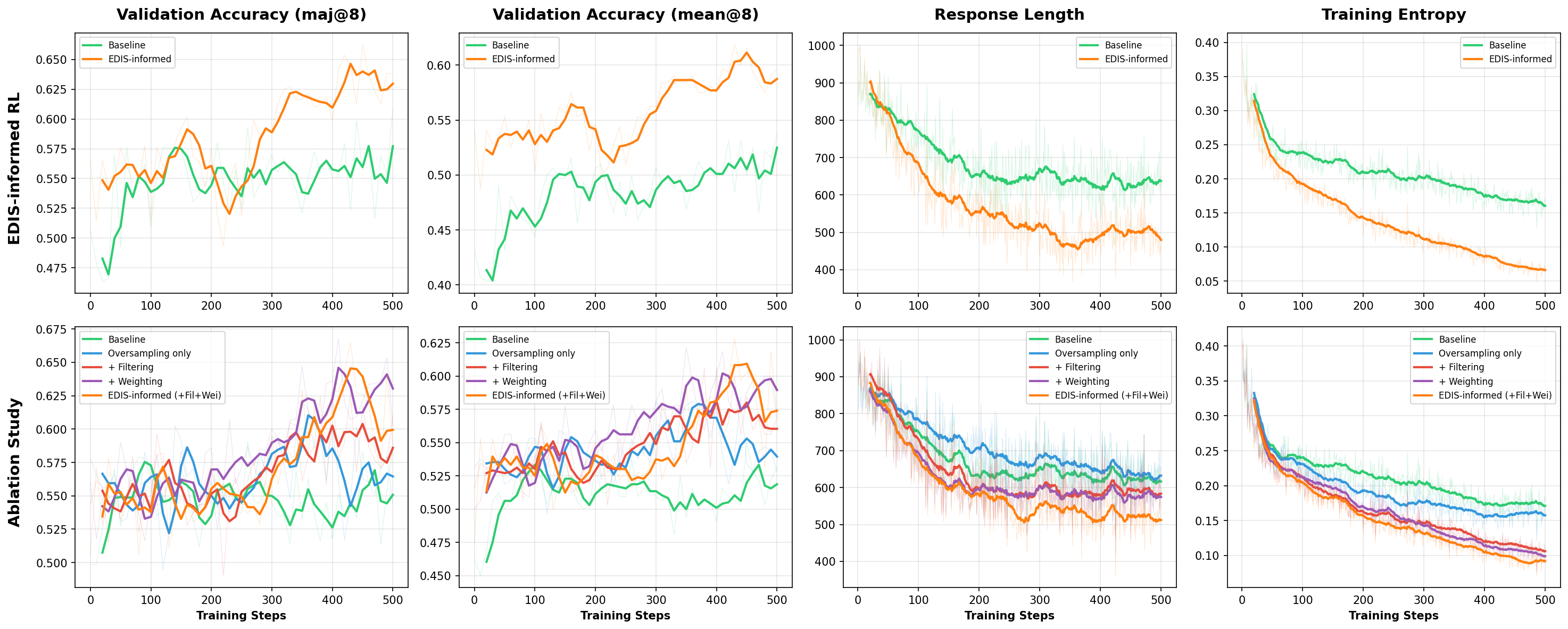}
\caption{Validating EDIS as a training signal. \textbf{Top row}: At temperature $T=0.6$, EDIS-informed training (filtering + weighting) vs.\ standard GRPO baseline. \textbf{Bottom row}: Complete ablation at $T=0.2$, isolating the contribution of each EDIS component.}
\label{fig:rl_main}
\end{figure*}

To examine whether EDIS can provide useful training signals for RL, we design experiments comparing training with and without EDIS guidance. We also investigate whether EDIS-based test-time selection remains beneficial as RL training proceeds. Our primary goal is to verify that EDIS meaningfully distinguishes informative trajectories---not to optimize RL performance.

\paragraph{Setting.} All experiments use Qwen2.5-Math-1.5B trained on NuminaMath-20K for 500 steps with GRPO ($k=8$ responses per prompt), validating on AMC23 every 10 steps. We conduct two experiment sets: (1) at $T=0.6$, we compare full EDIS-informed training against the baseline; (2) at $T=0.2$, we perform a complete ablation to isolate each component's contribution.

\paragraph{Configurations.} Five settings progressively incorporate EDIS: (1) \emph{Standard GRPO}: baseline without EDIS; (2) \emph{+ Oversampling}: adds test-time EDIS selection ($4\times$ candidates) without training-time signals; (3) \emph{+ EDIS filtering}: adds training-time filtering ($m=1.25$; Section~\ref{sec:edis_rl}), retaining only the most stable correct and most unstable incorrect responses; (4) \emph{+ EDIS weighting}: adds training-time weighting ($\alpha=1.8$), assigning differential importance via softmax over signed z-scores---lower $\alpha$ concentrates weight on extremes, higher $\alpha$ yields more uniform weighting; (5) \emph{Filtering + Weighting}: combines both mechanisms ($m=1.25$, $\alpha=1.8$)---the full \emph{EDIS-informed} configuration used at $T=0.6$. Configurations (2)--(5) share the same test-time EDIS selection; differences lie solely in training-time signals.

\begin{table}[h]
\centering
\caption{EDIS as a training signal (best validation accuracy). $T=0.6$: full EDIS-informed vs.\ baseline. $T=0.2$: component ablation.}
\label{tab:rl_summary}
\small
\setlength{\tabcolsep}{5pt}
\begin{tabular}{clcc}
\toprule
$T$ & Configuration & maj@8 & mean@8 \\
\midrule
\multirow{2}{*}{$0.6$}
& Standard GRPO & $60.8\%$ & $53.8\%$ \\
& EDIS-informed & $\mathbf{66.2\%}$ {\scriptsize($+5.4$)} & $\mathbf{61.9\%}$ {\scriptsize($+8.1$)} \\
\midrule
\multirow{5}{*}{$0.2$}
& Standard GRPO & $59.1\%$ & $55.0\%$ \\
& + Oversampling & $62.1\%$ {\scriptsize($+3.0$)} & $58.4\%$ {\scriptsize($+3.4$)} \\
& + EDIS filtering & $62.6\%$ {\scriptsize($+3.5$)} & $60.6\%$ {\scriptsize($+5.6$)} \\
& + EDIS weighting & $\mathbf{66.8\%}$ {\scriptsize($\mathbf{+7.7}$)} & $62.2\%$ {\scriptsize($+7.2$)} \\
& Filtering + Weighting & $66.5\%$ {\scriptsize($+7.4$)} & $\mathbf{62.8\%}$ {\scriptsize($\mathbf{+7.8}$)} \\
\bottomrule
\end{tabular}
\end{table}

\paragraph{Results.} Figure~\ref{fig:rl_main} and Table~\ref{tab:rl_summary} show consistent patterns across both temperatures. At $T=0.6$, full EDIS-informed training yields substantial gains over the baseline ($+5.4$ pp maj@8, $+8.1$ pp mean@8). The $T=0.2$ ablation isolates each component: test-time selection alone (oversampling) provides meaningful benefit ($+3.0$ pp maj@8), confirming that EDIS remains useful for selecting among RL-trained outputs. Training-time filtering adds modest additional signal ($+3.5$ pp total), while weighting achieves the largest gains ($+7.7$ pp maj@8). Combining both yields the best mean@8 ($+7.8$ pp). Beyond accuracy, EDIS-informed training produces dramatically lower entropy ($0.07$--$0.09$ vs.\ $0.16$--$0.18$) and shorter responses ($453$--$525$ vs.\ $620$--$646$ tokens). EDIS maintains stable discriminative power throughout training, with the spike ratio between incorrect and correct responses at $1.9$--$2.7\times$ across all 500 steps (Appendix~\ref{app:stability}; additional training dynamics in Appendix~\ref{app:edis_training}).

\paragraph{Interpretation.} These results connect directly to EDIS's core insight: entropy dynamics distinguish trajectories by \emph{how} reasoning unfolds, not just final correctness. The lower entropy and shorter responses reflect more focused reasoning, where the model avoids uncertainty cascades characteristic of incorrect trajectories. Notably, entropy reduction is more pronounced at $T=0.6$ ($59\%$ vs.\ $50\%$ at $T=0.2$), suggesting EDIS provides greater benefit when baseline generation exhibits higher variability. The gap between filtering ($+3.5$ pp) and weighting ($+7.7$ pp) reveals an important distinction: both use the same number of training samples, but filtering reshapes the EDIS distribution by retaining only extremes (low-EDIS correct, high-EDIS incorrect) and discarding ambiguous cases, while weighting preserves the full distribution with differential importance. This suggests that ``middle ground'' samples---moderately stable correct and moderately unstable incorrect responses---still carry useful gradient signal when appropriately weighted.

\section{Limitations and Future Work}
\label{sec:discussion}

\paragraph{Limitations.}
This investigation focuses on mathematical reasoning, where correctness is objectively verifiable. Whether the instability patterns transfer to other reasoning-intensive domains---code generation, scientific reasoning, logical deduction---remains to be validated. Our RL experiments serve as a proof-of-concept; more sophisticated integration strategies may yield larger gains.

A more fundamental limitation concerns EDIS's generality across models. While the qualitative patterns appear consistently, optimal thresholds and parameters vary across model families and sizes. Window sizes, rebound thresholds, and spike weightings require calibration for new models, as entropy dynamics depend on model-specific factors such as vocabulary size and training distribution.

\paragraph{Future Work.}
Two directions seem particularly promising. First, \emph{token-level credit assignment}: extending EDIS to identify which tokens contribute most to instability could enable fine-grained feedback for process reward models. Second, \emph{unsupervised process supervision}: extending trajectory-level signals to step-level analysis---by segmenting at reasoning boundaries and computing local instability---could help bootstrap process reward models without ground-truth step labels.

\section{Conclusion}
\label{sec:conclusion}

We introduced EDIS, a trajectory-level metric that captures instability patterns in entropy evolution during LLM reasoning. The central insight is that reasoning quality can be diagnosed from how confidence \emph{evolves} during generation, not just its average value. By shifting from static to dynamic analysis, EDIS extracts richer signal from token-level data that prior methods reduce to summary statistics. The characteristic instability patterns---burst spikes and peak-valley spikes---persist across models, temperatures, and training stages, suggesting they reflect fundamental properties of reasoning failure. EDIS achieves an $82\%$ relative accuracy improvement for inference-time selection and up to $+7.7$ percentage points gains for RL training, consistently outperforming alternative confidence measures.

\bibliography{references}
\bibliographystyle{icml2026}

%%%%%%%%%%%%%%%%%%%%%%%%%%%%%%%%%%%%%%%%%%%%%%%%%%%%%%%%%%%%%%%%%%%%%%%%%%%%%%%
%%%%%%%%%%%%%%%%%%%%%%%%%%%%%%%%%%%%%%%%%%%%%%%%%%%%%%%%%%%%%%%%%%%%%%%%%%%%%%%
% APPENDIX
%%%%%%%%%%%%%%%%%%%%%%%%%%%%%%%%%%%%%%%%%%%%%%%%%%%%%%%%%%%%%%%%%%%%%%%%%%%%%%%
%%%%%%%%%%%%%%%%%%%%%%%%%%%%%%%%%%%%%%%%%%%%%%%%%%%%%%%%%%%%%%%%%%%%%%%%%%%%%%%
\newpage
\appendix
\onecolumn

\section{Token-Level Visualization of Entropy Dynamics}
\label{appendix:token_heatmap}

\begin{figure}[!h]
    \centering
    \includegraphics[width=\textwidth]{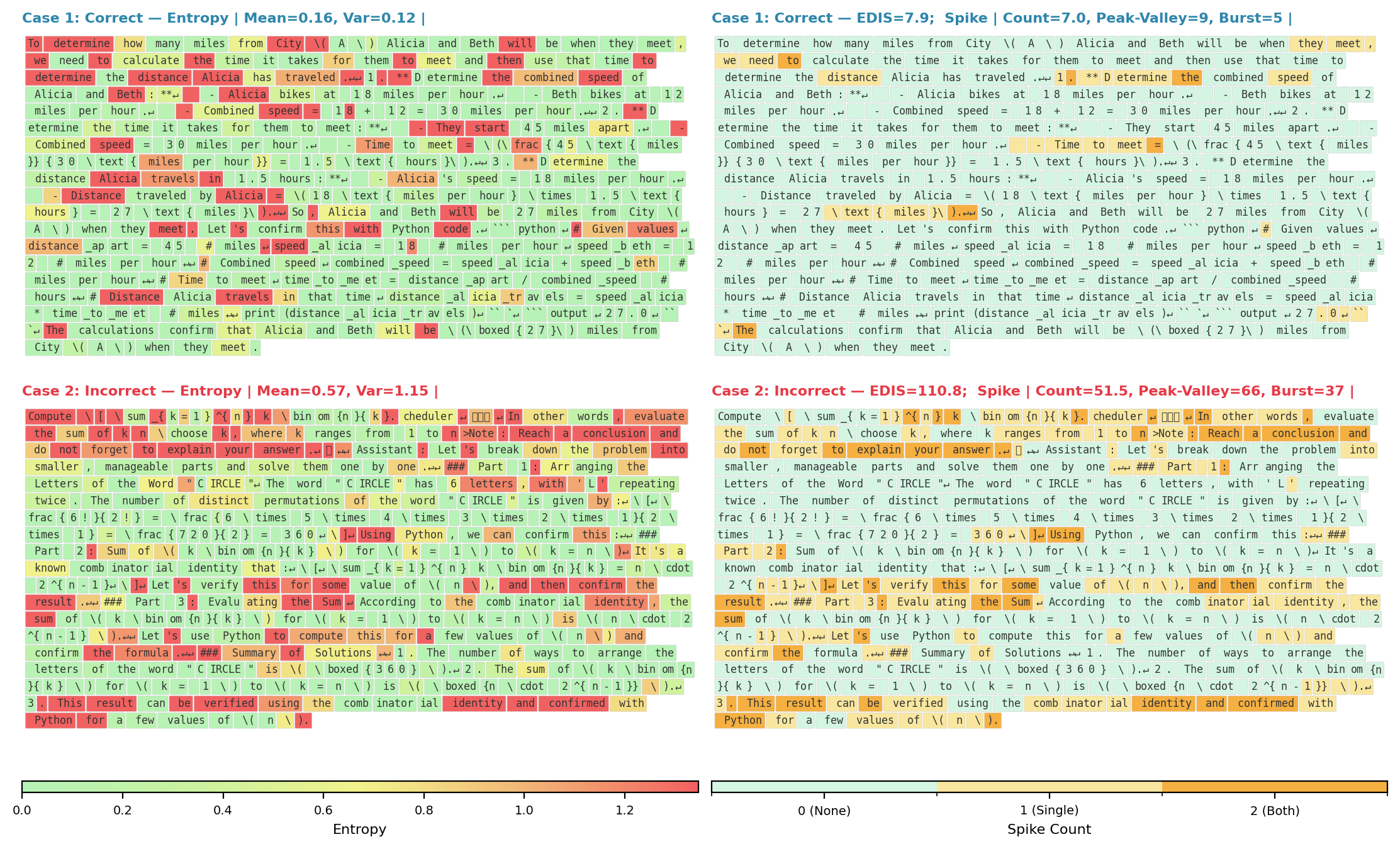}
    \caption{Token-level entropy and spike heatmaps. Left: entropy magnitude (green$\to$red). Right: spike status (green=none, yellow=single, orange=both). EDIS discrimination ratio ($14.0\times$) substantially exceeds mean entropy ($3.6\times$).}
    \label{fig:token_heatmap}
\end{figure}

Figure~\ref{fig:token_heatmap} provides a token-level view of the entropy dynamics illustrated in Figure~\ref{fig:entropy_patterns}. Each token is rendered with background color encoding either entropy magnitude (left columns) or spike status (right columns).

\paragraph{Entropy Heatmap.} Color ranges from green (low entropy) through yellow to red (high entropy), visualizing the model's confidence at each generation step.

\paragraph{Spike Heatmap.} Light green indicates no spike; yellow indicates a single spike type (burst or peak-valley); orange indicates both spike types co-occur at that position.

\paragraph{Qualitative Analysis.}
Case 1 (correct reasoning) exhibits predominantly green tokens with sparse, isolated perturbations. Case 2 (incorrect reasoning) shows extensive yellow/red regions in the entropy heatmap and pervasive yellow/orange regions in the spike heatmap. Notably, hallucinated tokens (e.g., ``cheduler'', Thai characters) concentrate in high-entropy, high-spike regions.

\paragraph{Discrimination Ratio.}
EDIS achieves a $14.0\times$ discrimination ratio between the two cases ($110.8$ vs.\ $7.9$), compared to only $3.6\times$ for mean entropy ($0.57$ vs.\ $0.16$). This nearly four-fold improvement stems from two mechanisms. First, the multiplicative formulation (Eq.~\ref{eq:edis}) compounds spike count and variance into an amplified signal. Second, the spike detection thresholds ($\tau_b$, $\tau_r$) act as a denoising filter: unlike raw entropy which fluctuates at every token, spikes are triggered only by \emph{significant} instability events---sustained entropy growth or sharp rebounds from historical minima. This sparsification yields cleaner separation: in Case 1, spikes appear infrequently (7.0 total) against a stable background, while in Case 2, spikes cluster densely (51.5 total) around failure regions. The result is a diagnostic signal that is both stronger in magnitude and less cluttered by noise.

\section{Statistical Analysis of Entropy Spikes}
\label{app:spike_analysis}

While EDIS uses sophisticated burst and peak-valley detection (Appendix~\ref{appendix:token_heatmap}), we also validate that a simpler spike definition yields statistically significant differences. We define an \emph{entropy spike} as a generation step where entropy changes abruptly: $|H_{t+1} - H_t| > \tau$ for a fixed threshold $\tau = 0.7$.

\begin{table}[h]
\centering
\caption{Entropy spike counts (threshold $\tau = 0.7$) for correct and incorrect solutions. Qwen2.5-Math-1.5B results are from RL training validation; Qwen3-4B-Instruct results are from multi-temperature evaluation.}
\label{tab:spike_counts}
\small
\begin{tabular}{lccccc}
\toprule
Model & Correct & Incorrect & Ratio & $p$-value & Cohen's $d$ \\
\midrule
Qwen2.5-Math-1.5B & $49.3 \pm 23.2$ & $82.0 \pm 39.3$ & $1.66\times$ & $< 10^{-100}$ & $1.03$ \\
Qwen3-4B-Instruct & $36.9 \pm 52.8$ & $133.8 \pm 145.6$ & $3.62\times$ & $< 10^{-100}$ & $0.97$ \\
\bottomrule
\end{tabular}
\end{table}

Incorrect solutions exhibit $1.7$--$3.6\times$ more entropy spikes than correct ones. This difference is statistically robust (Cohen's $d \approx 1.0$, $p < 10^{-100}$), indicating a large and consistent effect. EDIS's refined dual-threshold detection further amplifies this separation to $14.0\times$ (Appendix~\ref{appendix:token_heatmap}), demonstrating the value of capturing sustained entropy growth and sharp rebounds beyond simple differencing.

\section{Detailed Best-of-$N$ Scaling Results}
\label{app:best_of_n_detail}

This section presents detailed EDIS-based best-$k$-of-$N$ selection results for three models, complementing the summary in Section~\ref{sec:edis_scaling}. Each figure shows performance across four benchmarks (rows) and three temperatures (columns), with lines indicating average accuracy, EDIS-best, and majority voting.

\subsection{Qwen2.5-Math-1.5B}

\begin{figure*}[!h]
  \centering
  \includegraphics[width=0.9\textwidth]{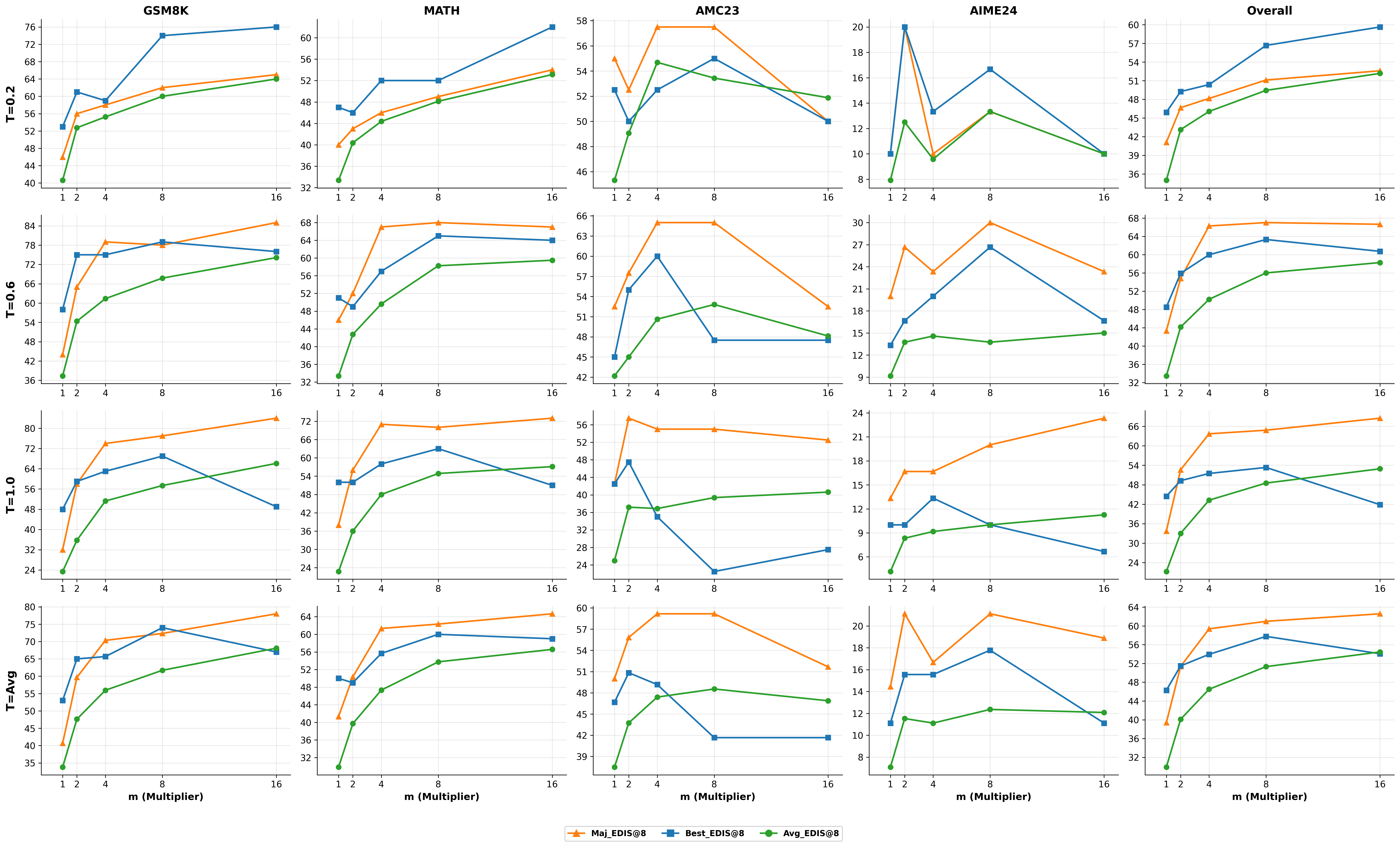}
  \caption{Best-$k$-of-$N$ selection for Qwen2.5-Math-1.5B. EDIS filtering more than doubles baseline accuracy on GSM8K ($33.8\% \to 68.1\%$) and MATH ($29.8\% \to 56.6\%$).}
  \label{fig:best_of_n_1.5b}
\end{figure*}

Figure~\ref{fig:best_of_n_1.5b} shows results for the smallest model. EDIS yields dramatic improvements: GSM8K average accuracy increases from $33.8\%$ to $68.1\%$, and majority voting reaches $85\%$ at temperature $0.6$. Even at high temperature ($1.0$), where outputs are noisiest, majority voting improves from $32\%$ to $84\%$ on GSM8K.

\subsection{Qwen3-4B-Instruct}

\begin{figure*}[!h]
  \centering
  \includegraphics[width=0.9\textwidth]{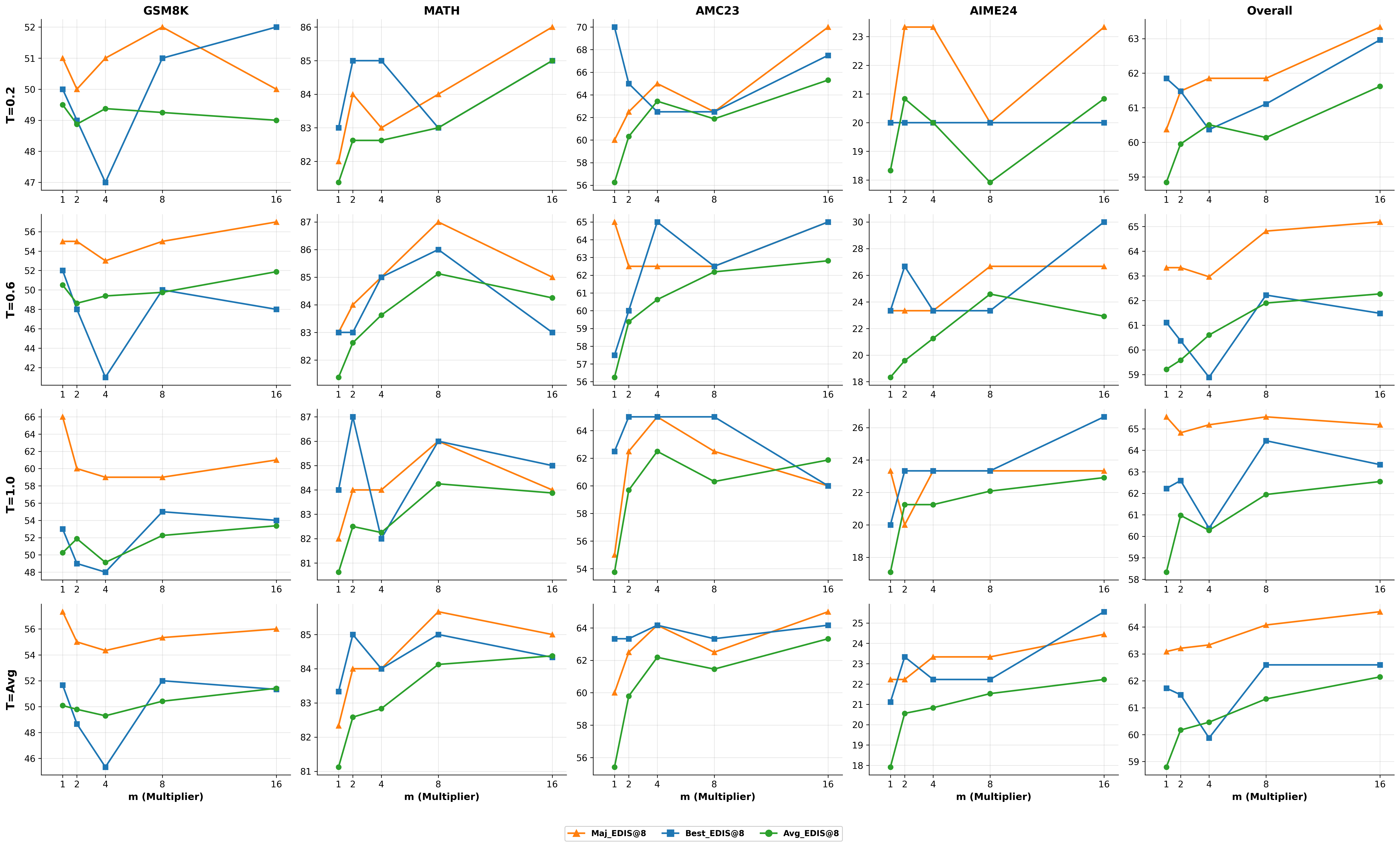}
  \caption{Best-$k$-of-$N$ selection for Qwen3-4B-Instruct. Despite high baseline performance ($>80\%$ on MATH), EDIS provides consistent gains on competition benchmarks.}
  \label{fig:best_of_n_qwen3}
\end{figure*}

Figure~\ref{fig:best_of_n_qwen3} shows results for the instruction-tuned model with the highest baseline. Overall accuracy improves from $58.8\%$ to $62.2\%$ (+$3.4$ pp). While MATH is near-saturated ($81$--$85\%$), EDIS provides meaningful gains on competition benchmarks: AMC23 improves by $7.9$ pp and AIME24 by $4.3$ pp.

\subsection{Qwen2.5-Math-7B}

\begin{figure*}[!h]
  \centering
  \includegraphics[width=0.9\textwidth]{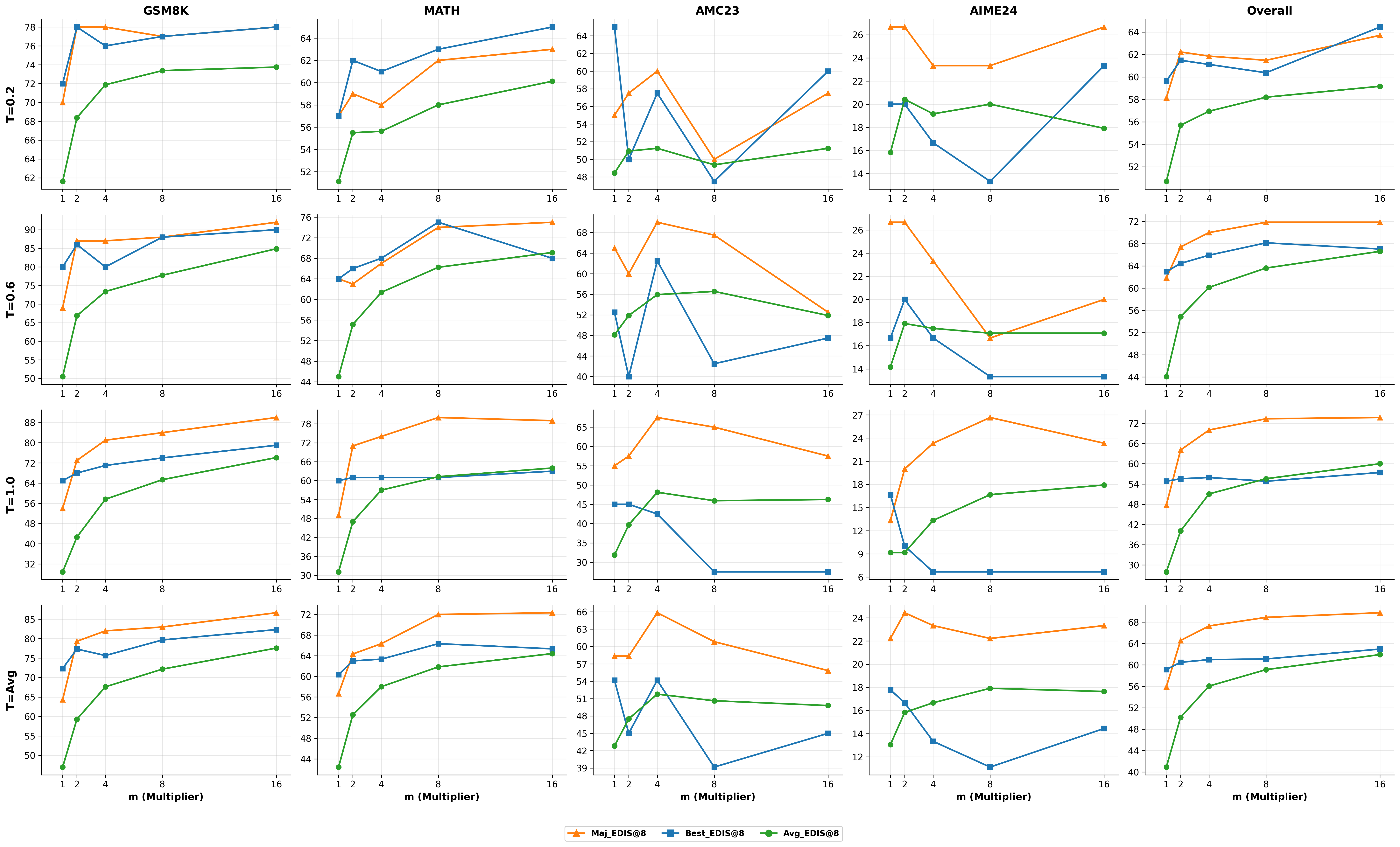}
  \caption{Best-$k$-of-$N$ selection for Qwen2.5-Math-7B. EDIS filtering yields $+21$ pp overall improvement and enables $92\%$ majority voting accuracy on GSM8K at temperature $0.6$.}
  \label{fig:best_of_n_qwen7b}
\end{figure*}

Figure~\ref{fig:best_of_n_qwen7b} shows results for the 7B model. Overall average accuracy improves from $40.9\%$ to $61.9\%$ (+$21.0$ pp), with majority voting reaching $69.8\%$. GSM8K shows the strongest gains: majority voting achieves $92\%$ at temperature $0.6$. At high temperature, EDIS recovers performance from $28.0\%$ to $60.0\%$, demonstrating robustness to noisy sampling.

\subsection{Summary}

Across all three models, EDIS-based filtering consistently improves accuracy as candidate pool size increases. The benefits scale inversely with model capability: Qwen2.5-Math-1.5B gains $+24.6$ pp, Qwen2.5-Math-7B gains $+21.0$ pp, and Qwen3-4B-Instruct gains $+3.4$ pp. Importantly, EDIS remains effective even for strong models, providing meaningful improvements on challenging competition benchmarks where room for gains exists. These results confirm that entropy trajectory stability is a robust indicator of reasoning quality across model sizes, temperatures, and problem difficulties.

\section{Full Scaling Comparison Results}
\label{app:full_scaling}

This section presents complete results for the comparison of selection methods (Section~\ref{sec:edis_comparison}), broken down by sampling temperature. Table~\ref{tab:scaling_all} reports accuracy for each method at temperatures $0.2$, $0.6$, and $1.0$. All experiments use Qwen2.5-Math-1.5B.

\begin{table*}[!h]
\centering
\caption{Selection method comparison across all temperatures. Best results per cell group in \textbf{bold}. Maj = Majority Voting, Ent = Sequence Entropy, SC = Self-Certainty.}
\label{tab:scaling_all}
\resizebox{\textwidth}{!}{%
\renewcommand{\arraystretch}{1.3}
\setlength{\tabcolsep}{3.8pt}
\scriptsize
\begin{tabular}{cl|ccccc|ccccc|ccccc|ccccc}
\toprule
& & \multicolumn{5}{c|}{Temperature 0.2} & \multicolumn{5}{c|}{Temperature 0.6} & \multicolumn{5}{c|}{Temperature 1.0} & \multicolumn{5}{c}{Average} \\
$N$ & Dataset & Rand & Maj & Ent & SC & EDIS & Rand & Maj & Ent & SC & EDIS & Rand & Maj & Ent & SC & EDIS & Rand & Maj & Ent & SC & EDIS \\
\midrule
\multirow{5}{*}{1}
 & GSM8K & 44.38 & 50.00 & 46.00 & 51.00 & \textbf{59.00} & 40.12 & 57.00 & 54.00 & 63.00 & \textbf{69.00} & 23.88 & 36.00 & \textbf{56.00} & 52.00 & \textbf{56.00} & 36.12 & 47.67 & 52.00 & 55.33 & \textbf{61.33} \\
 & MATH & 33.00 & 39.00 & 38.00 & 42.00 & \textbf{48.00} & 33.00 & 49.00 & 45.00 & 51.00 & \textbf{53.00} & 25.12 & 44.00 & 53.00 & \textbf{54.00} & 49.00 & 30.38 & 44.00 & 45.33 & 49.00 & \textbf{50.00} \\
 & AMC23 & 49.38 & \textbf{60.00} & 57.50 & 57.50 & \textbf{60.00} & 41.88 & 62.50 & \textbf{65.00} & 62.50 & 62.50 & 23.44 & 40.00 & 40.00 & 37.50 & \textbf{45.00} & 38.23 & 54.17 & 54.17 & 52.50 & \textbf{55.83} \\
 & AIME24 & 8.75 & 13.33 & 16.67 & 16.67 & \textbf{20.00} & 8.33 & 16.67 & 16.67 & 16.67 & \textbf{20.00} & 3.33 & 6.67 & 6.67 & \textbf{10.00} & 3.33 & 6.81 & 12.22 & 13.33 & \textbf{14.44} & \textbf{14.44} \\
 & \textit{Overall} & 36.94 & 43.33 & 41.48 & 44.81 & \textbf{50.74} & 34.21 & 50.37 & 48.15 & 53.33 & \textbf{56.67} & 21.99 & 36.30 & \textbf{47.04} & 45.93 & 45.93 & 31.05 & 43.33 & 45.56 & 48.02 & \textbf{51.11} \\
\midrule
\multirow{5}{*}{2}
 & GSM8K & 43.19 & 48.00 & 44.00 & 51.00 & \textbf{57.00} & 39.62 & 60.00 & 50.00 & 63.00 & \textbf{73.00} & 23.88 & 43.00 & 65.00 & 57.00 & \textbf{69.00} & 35.56 & 50.33 & 53.00 & 57.00 & \textbf{66.33} \\
 & MATH & 32.25 & 43.00 & 39.00 & 43.00 & \textbf{46.00} & 31.25 & 45.00 & 43.00 & 48.00 & \textbf{61.00} & 20.62 & 39.00 & 59.00 & 57.00 & \textbf{60.00} & 28.04 & 42.33 & 47.00 & 49.33 & \textbf{55.67} \\
 & AMC23 & 47.34 & \textbf{60.00} & 57.50 & \textbf{60.00} & \textbf{60.00} & 42.50 & \textbf{60.00} & \textbf{60.00} & \textbf{60.00} & 57.50 & 25.94 & 52.50 & \textbf{65.00} & 60.00 & 45.00 & 38.59 & 57.50 & \textbf{60.83} & 60.00 & 54.17 \\
 & AIME24 & 8.54 & 10.00 & 10.00 & 10.00 & \textbf{13.33} & 7.50 & \textbf{20.00} & \textbf{20.00} & \textbf{20.00} & \textbf{20.00} & 4.17 & \textbf{16.67} & \textbf{16.67} & 13.33 & 13.33 & 6.74 & \textbf{15.56} & \textbf{15.56} & 14.44 & \textbf{15.56} \\
 & \textit{Overall} & 35.90 & 43.70 & 40.37 & 44.81 & \textbf{48.52} & 33.38 & 50.00 & 45.56 & 52.22 & \textbf{60.37} & 20.79 & 40.00 & \textbf{57.41} & 52.59 & 55.93 & 30.02 & 44.57 & 47.78 & 49.88 & \textbf{54.94} \\
\midrule
\multirow{5}{*}{4}
 & GSM8K & 44.59 & 49.00 & 45.00 & 51.00 & \textbf{58.00} & 40.19 & 61.00 & 47.00 & 61.00 & \textbf{74.00} & 23.19 & 32.00 & 68.00 & 57.00 & \textbf{70.00} & 35.99 & 47.33 & 53.33 & 56.33 & \textbf{67.33} \\
 & MATH & 33.59 & 41.00 & 39.00 & 42.00 & \textbf{49.00} & 32.91 & 51.00 & 49.00 & 56.00 & \textbf{61.00} & 23.38 & 47.00 & \textbf{64.00} & 61.00 & \textbf{64.00} & 29.96 & 46.33 & 50.67 & 53.00 & \textbf{58.00} \\
 & AMC23 & 48.75 & \textbf{60.00} & 57.50 & \textbf{60.00} & 57.50 & 40.55 & \textbf{60.00} & \textbf{60.00} & \textbf{60.00} & 57.50 & 27.27 & 50.00 & \textbf{57.50} & 50.00 & 52.50 & 38.85 & 56.67 & \textbf{58.33} & 56.67 & 55.83 \\
 & AIME24 & 8.75 & 13.33 & 13.33 & 13.33 & \textbf{16.67} & 7.81 & 20.00 & 20.00 & 16.67 & \textbf{23.33} & 4.90 & 16.67 & \textbf{23.33} & 20.00 & \textbf{23.33} & 7.15 & 16.67 & 18.89 & 16.67 & \textbf{21.11} \\
 & \textit{Overall} & 37.15 & 43.70 & 41.11 & 44.81 & \textbf{50.00} & 33.95 & 52.59 & 46.67 & 54.07 & \textbf{61.11} & 21.83 & 38.52 & \textbf{60.00} & 53.33 & \textbf{60.00} & 30.98 & 44.94 & 49.26 & 50.74 & \textbf{57.04} \\
\midrule
\multirow{5}{*}{8}
 & GSM8K & 44.47 & 51.00 & 49.00 & 55.00 & \textbf{65.00} & 41.44 & 63.00 & 55.00 & 70.00 & \textbf{76.00} & 22.98 & 35.00 & 66.00 & 51.00 & \textbf{76.00} & 36.30 & 49.67 & 56.67 & 58.67 & \textbf{72.33} \\
 & MATH & 33.77 & 42.00 & 40.00 & 43.00 & \textbf{47.00} & 31.27 & 49.00 & 47.00 & 55.00 & \textbf{68.00} & 23.36 & 47.00 & 64.00 & 59.00 & \textbf{65.00} & 29.46 & 46.00 & 50.33 & 52.33 & \textbf{60.00} \\
 & AMC23 & 47.70 & \textbf{57.50} & \textbf{57.50} & \textbf{57.50} & \textbf{57.50} & 41.37 & \textbf{62.50} & \textbf{62.50} & \textbf{62.50} & \textbf{62.50} & 25.43 & 57.50 & \textbf{60.00} & 57.50 & 52.50 & 38.16 & 59.17 & \textbf{60.00} & 59.17 & 57.50 \\
 & AIME24 & 9.17 & 10.00 & 10.00 & 10.00 & \textbf{13.33} & 8.59 & \textbf{20.00} & \textbf{20.00} & \textbf{20.00} & \textbf{20.00} & 4.79 & 20.00 & 20.00 & \textbf{23.33} & 20.00 & 7.52 & 16.67 & 16.67 & \textbf{17.78} & \textbf{17.78} \\
 & \textit{Overall} & 37.06 & 44.07 & 42.59 & 45.93 & \textbf{51.48} & 34.01 & 52.96 & 49.26 & 57.78 & \textbf{64.81} & 21.46 & 41.11 & 59.26 & 51.85 & \textbf{62.22} & 30.84 & 46.05 & 50.37 & 51.85 & \textbf{59.51} \\
\midrule
\multirow{5}{*}{16}
 & GSM8K & 44.04 & 50.00 & 46.00 & 52.00 & \textbf{59.00} & 40.55 & 64.00 & 52.00 & 68.00 & \textbf{81.00} & 22.07 & 33.00 & 67.00 & 48.00 & \textbf{77.00} & 35.55 & 49.00 & 55.00 & 56.00 & \textbf{72.33} \\
 & MATH & 33.48 & 42.00 & 40.00 & 43.00 & \textbf{51.00} & 32.41 & 54.00 & 50.00 & 60.00 & \textbf{71.00} & 22.65 & 46.00 & \textbf{66.00} & 59.00 & 65.00 & 29.51 & 47.33 & 52.00 & 54.00 & \textbf{62.33} \\
 & AMC23 & 47.42 & 55.00 & \textbf{57.50} & 55.00 & \textbf{57.50} & 41.07 & \textbf{62.50} & \textbf{62.50} & \textbf{62.50} & 57.50 & 24.22 & 55.00 & \textbf{62.50} & \textbf{62.50} & 52.50 & 37.57 & 57.50 & \textbf{60.83} & 60.00 & 55.83 \\
 & AIME24 & 8.44 & \textbf{13.33} & \textbf{13.33} & \textbf{13.33} & \textbf{13.33} & 7.89 & 23.33 & 23.33 & 23.33 & \textbf{26.67} & 4.48 & 16.67 & 23.33 & 20.00 & \textbf{26.67} & 6.94 & 17.78 & 20.00 & 18.89 & \textbf{22.22} \\
 & \textit{Overall} & 36.67 & 43.70 & 41.85 & 44.81 & \textbf{50.74} & 33.98 & 55.56 & 49.63 & 59.26 & \textbf{67.78} & 20.65 & 39.26 & 61.11 & 51.11 & \textbf{63.33} & 30.43 & 46.17 & 50.86 & 51.73 & \textbf{60.62} \\
\bottomrule
\end{tabular}%
}
\end{table*}

\paragraph{Analysis by Temperature.}
At low temperature ($0.2$), the model generates more deterministic outputs, resulting in lower variance across methods. EDIS maintains a consistent advantage, particularly on GSM8K where it achieves $57$--$65\%$ accuracy compared to $44$--$49\%$ for sequence entropy.
At moderate temperature ($0.6$), EDIS shows its strongest advantage, achieving up to $81\%$ on GSM8K at $N=16$ compared to $68\%$ for self-certainty---a gap of $13$ percentage points.
At high temperature ($1.0$), outputs are most diverse but also noisiest. Sequence entropy becomes more competitive, matching or exceeding EDIS in $15$ out of $25$ dataset-$N$ combinations. However, EDIS maintains the best overall performance on GSM8K ($56$--$77\%$) and achieves the highest accuracy at larger candidate pools ($N \geq 4$).

\paragraph{Summary.}
Across all temperatures and candidate pool sizes, EDIS achieves the best or near-best overall accuracy. The advantage is most pronounced at moderate temperature ($0.6$), where output diversity is balanced with quality. At high temperature ($1.0$), the gap between methods narrows, but EDIS remains competitive. These results demonstrate that EDIS is robust across different sampling configurations.

\section{Stability of EDIS Across Training}
\label{app:stability}

A key question for practical deployment is whether EDIS requires recalibration as training progresses. If the discriminative power of entropy dynamics varied significantly across checkpoints, practitioners would need to tune thresholds or validate performance at each stage. We analyze this question by tracking spike ratios throughout RL training on Qwen2.5-Math-1.5B (500 steps on NuminaMath-20K), evaluating on the AMC23 validation set every 10 steps.

\begin{figure}[h]
\centering
\includegraphics[width=\textwidth]{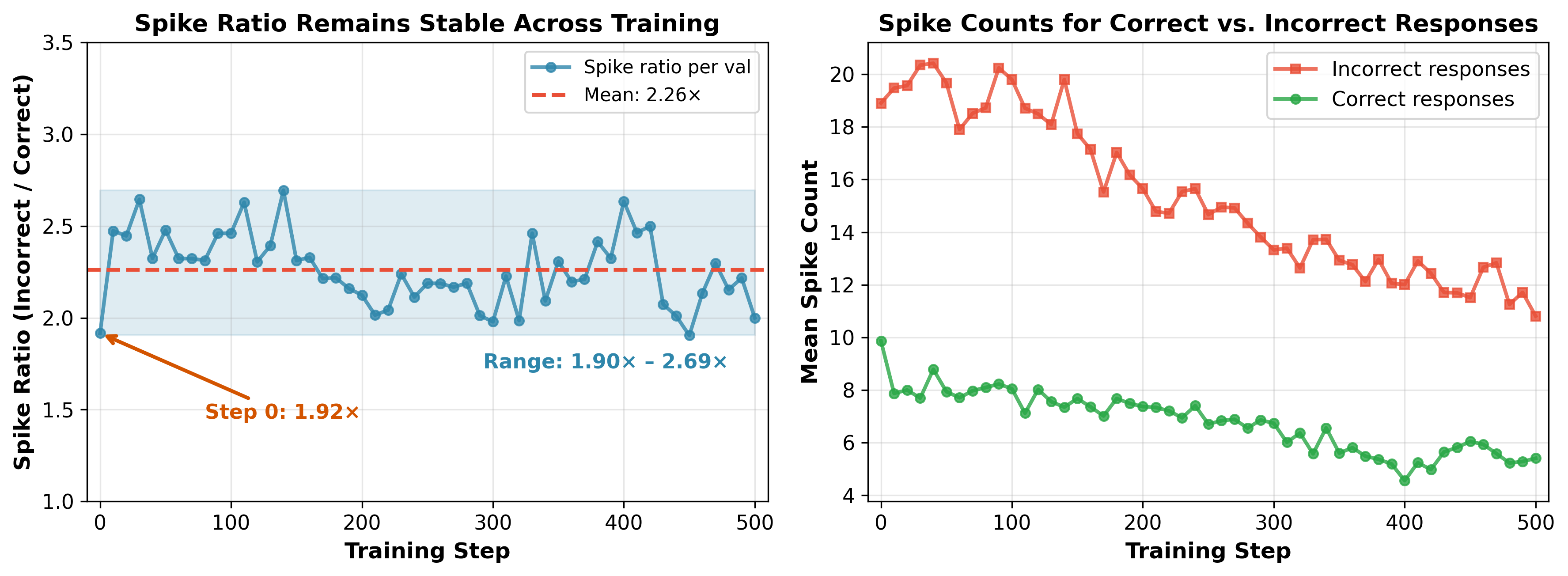}
\caption{Spike ratio stability across 500 training steps (evaluated on AMC23). \textbf{Left}: The ratio of spike counts (incorrect/correct) remains stable at $1.9$--$2.7\times$ throughout training, with Step 0 already showing strong discriminative power ($1.92\times$). \textbf{Right}: While absolute spike counts decrease for both correct and incorrect responses as training progresses, the relative difference persists.}
\label{fig:spike_ratio_training}
\end{figure}

\paragraph{Spike ratio remains stable.}
Figure~\ref{fig:spike_ratio_training} (left) shows the ratio of mean spike counts between incorrect and correct responses at each training checkpoint. At Step 0 (the pretrained checkpoint, before any RL fine-tuning), incorrect responses already exhibit $1.92\times$ more entropy spikes than correct ones. This ratio remains remarkably stable throughout training, fluctuating between $1.90\times$ and $2.69\times$ with a mean of $2.26\times$ and standard deviation of only $0.19$.

\paragraph{Absolute counts decrease, relative difference persists.}
Figure~\ref{fig:spike_ratio_training} (right) reveals an interesting pattern: both correct and incorrect responses show decreasing spike counts as training progresses, reflecting the model's increasing confidence. However, the \emph{gap} between them remains consistent---incorrect responses consistently exhibit more instability than correct ones, regardless of the overall confidence level.

\paragraph{Implication for deployment.}
This stability has significant practical implications:
\begin{itemize}
    \item \textbf{No checkpoint-specific calibration required}: The same spike detection thresholds ($\tau_b = 1.36$, $\tau_r = 1.33$) that work at Step 0 remain effective at Step 500.
    \item \textbf{Applicable to pretrained models}: EDIS provides useful signal even before any task-specific fine-tuning, enabling immediate deployment.
    \item \textbf{Robust across training dynamics}: The discriminative power of EDIS is not an artifact of a particular training stage but reflects intrinsic properties of reasoning quality.
\end{itemize}

These findings support the use of EDIS as a general-purpose diagnostic signal that does not require task-specific or checkpoint-specific tuning.

\section{Supplementary Details for RL Training}
\label{app:edis_training}

\begin{figure}[!h]
\centering
\includegraphics[width=\textwidth]{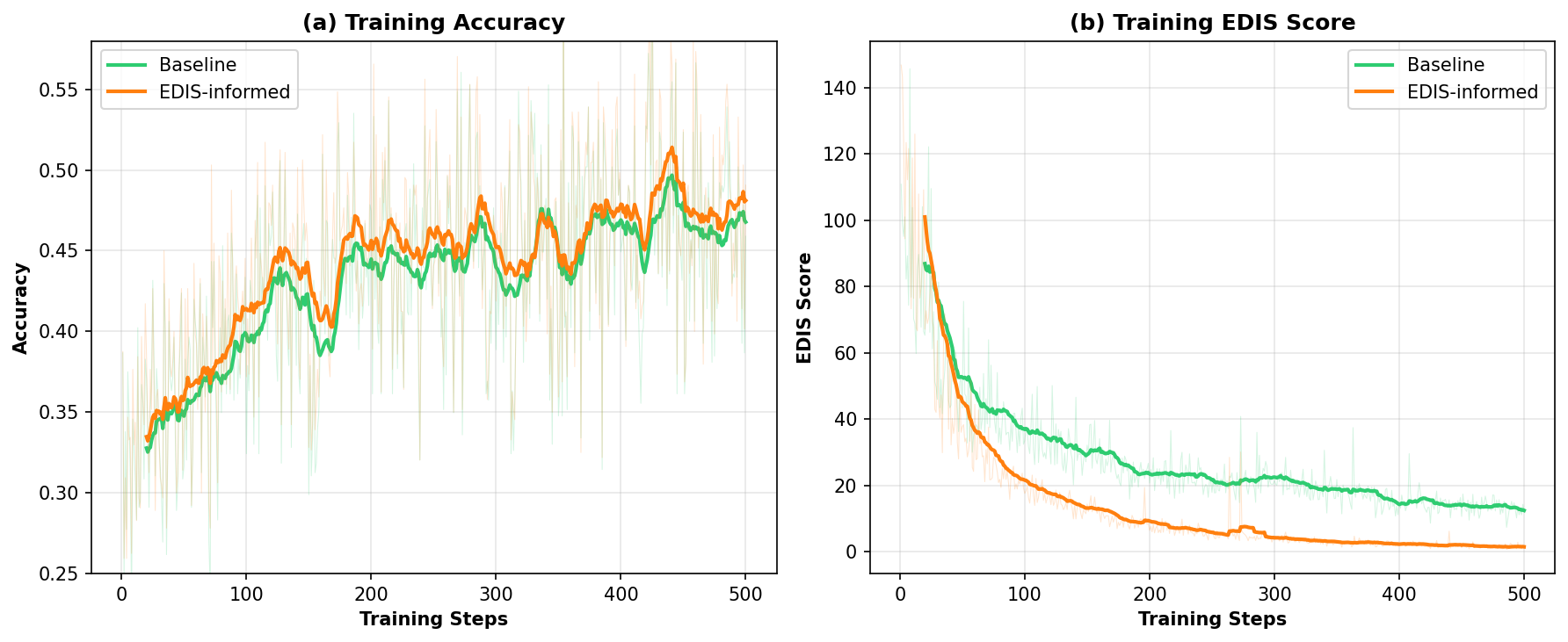}
\caption{Training dynamics at temperature $0.6$. \textbf{(a)} Accuracy improves throughout training; EDIS-informed achieves higher peak. \textbf{(b)} EDIS scores decrease as models learn; EDIS-informed reaches near-zero instability.}
\label{fig:edis_training}
\end{figure}

Figure~\ref{fig:edis_training} provides supplementary training curves for Section~\ref{sec:edis_rl_exp} at temperature $T=0.6$.

\paragraph{Accuracy.}
Training improves accuracy from $\sim$33\% to $\sim$48--52\% over 500 steps. Notably, EDIS-informed training achieves a slightly higher peak ($52.0\%$ vs.\ $48.0\%$), suggesting that optimizing for entropy stability complements rather than competes with the correctness objective.

\paragraph{EDIS Score.}
As expected, EDIS decreases throughout training for both methods—models naturally become more stable as they improve. The key difference is magnitude: our approach reduces EDIS to near-zero ($<5$), while baseline plateaus around $\sim$15. This persistent gap indicates that standard training leaves residual instability that explicit EDIS-based curation can eliminate.

\paragraph{Convergence.}
An interesting pattern emerges in how the two methods converge. Baseline shows rapid early gains but levels off mid-training, whereas EDIS-informed training continues improving steadily. This suggests that EDIS provides a useful learning signal even after task accuracy has saturated.

%%%%%%%%%%%%%%%%%%%%%%%%%%%%%%%%%%%%%%%%%%%%%%%%%%%%%%%%%%%%%%%%%%%%%%%%%%%%%%%
%%%%%%%%%%%%%%%%%%%%%%%%%%%%%%%%%%%%%%%%%%%%%%%%%%%%%%%%%%%%%%%%%%%%%%%%%%%%%%%

\end{document}